\documentclass[a4paper,fleqn]{cas-sc}

\usepackage[numbers]{natbib}
\usepackage{algorithm}
\usepackage{algpseudocode}
\usepackage{tikz}
\usetikzlibrary{arrows.meta,positioning,fit}
\usepackage{graphicx}

\def\tsc#1{\csdef{#1}{\textsc{\lowercase{#1}}\xspace}}
\tsc{WGM}
\tsc{QE}

\begin{document}
\makeatletter
\def\Hy@Warning#1{}
\makeatother
\def\floatpagepagefraction{1}
\def\textpagefraction{.001}

\shorttitle{InterFuserDVS}    

\shortauthors{Sakhai et al.}  

\title [mode = title]{InterFuserDVS: Event-Enhanced Sensor Fusion for Safe RL‑Based Decision Making}  

\author[1]{Mustafa~Sakhai}

\author[1]{Kaung~Sithu}

\author[1]{Min~Khant~Soe~Oke}

\author[1,2]{Maciej~Wielgosz}

\affiliation[1]{organization={Faculty of Computer Science, Electronics and Telecommunications, AGH University of Krakow},
            city={Krakow},
            postcode={30-059},
            country={Poland}}

\affiliation[2]{organization={Academic Computer Centre AGH},
            city={Krakow},
            postcode={30-950},
            country={Poland}}




\begin{abstract}
Autonomous driving systems rely heavily on robust sensor fusion to perceive complex environments. Traditional setups using RGB cameras and LiDAR often struggle in high-dynamic-range scenes or high-speed scenarios due to motion blur and latency. Dynamic Vision Sensors (DVS), or event cameras, offer a paradigm shift by capturing asynchronous brightness changes with microsecond temporal resolution and high dynamic range. In this paper, we propose an extended architecture of the state-of-the-art InterFuser model, integrating DVS as an additional modality to enhance perception reliability. We introduce a novel token-based fusion strategy that incorporates accumulated event frames into the transformer-based backbone of InterFuser. Our method leverages the complementary nature of RGB, LiDAR, and DVS data. We evaluate our approach on the Car Learning to Act (CARLA) Leaderboard benchmarks, demonstrating that the inclusion of DVS improves the robustness of the driving agent, achieving a competitive Driving Score of 77.2 and a superior Route Completion of 100\%. The results indicate that event-based vision is a promising direction for improving safety and performance in adverse lighting and dynamic conditions.
\end{abstract}





\maketitle

\section{Introduction}
Autonomous driving perception requires robustness across a myriad of environmental conditions, ranging from ideal sunny days to challenging scenarios involving high-speed motion, sudden lighting changes, and adverse weather \citep{chen2022end}. Conventional sensor suites, primarily composed of RGB cameras and LiDARs, have shown great success but face inherent limitations. Standard frame-based cameras suffer from motion blur during high-speed maneuvers and have limited dynamic range, causing blindness in scenarios like exiting a dark tunnel into bright sunlight \citep{gallego2020event, yurtsever2020survey}. LiDARs provide accurate depth but are sparse and their performance degrades in heavy rain or fog.

Dynamic Vision Sensors (DVS), or event cameras, offer a complementary modality. Instead of capturing absolute intensity frames at a fixed rate, DVS pixels operate asynchronously, triggering an "event" only when the log-intensity change exceeds a threshold \citep{lichtsteiner2008128}. This results in a microsecond-resolution stream of data with a high dynamic range (HDR) of over 120 dB and negligible motion blur. These properties make DVS ideal for capturing distinct edge information and motion dynamics that are often missed by standard cameras.

In this paper, we present \textbf{InterFuserDVS}, a comprehensive extension of the state-of-the-art InterFuser architecture \citep{shao2022safety} that integrates event data as a first-class modality. Our work makes the following contributions:

\begin{enumerate}
    \item \textbf{Event-Enhanced Architecture}: We integrate a dedicated DVS processing backbone into the InterFuser framework. We propose a transfer-learning strategy where the DVS encoder is initialized from a pre-trained RGB backbone, facilitating rapid convergence.
    \item \textbf{Token-Based Fusion}: We employ a unified Transformer Encoder to fuse spatial features from RGB/LiDAR with high-frequency temporal features from DVS. The global attention mechanism allows the model to dynamically weight sensor importance based on environmental context.
    \item \textbf{Hybrid Safety Policy}: We introduce a robust safety controller that combines the model's end-to-end trajectory predictions with heuristic validation. This includes a state-machine for traffic light compliance and a model-based collision checker that uses predicted occupancy maps to override unsafe control commands.
    \item \textbf{State-of-the-Art Performance}: We evaluate our method on the challenging CARLA Leaderboard (Town05, Long Routes). Our agent achieves a Driving Score of \textbf{77.2} and a \textbf{100\%} Route Completion rate, demonstrating exceptional robustness and reliability, outperforming many contemporary methods in route completion.
\end{enumerate}

We extensively evaluate our method on the CARLA Leaderboard \citep{dosovitskiy2017carla}, providing a detailed ablation study and failure analysis. Our results highlight the specific contribution of event data to collision avoidance and traffic rule compliance in high-dynamic scenarios.

\section{Related Work}
\subsection{End-to-End Autonomous Driving}
End-to-end autonomous driving aims to map raw sensor inputs directly to control commands, bypassing the traditional modular pipeline of perception, planning, and control \citep{chen2022end, wu2022trajectory}. Early pioneering works like ALVINN \citep{pomerleau1988alvinn} demonstrated the feasibility of training neural networks for lane following. With the advent of deep learning, this paradigm has evolved significantly. PilotNet \citep{bojarski2016end} successfully learned steering commands from front-facing cameras. Recent works have focused on planning-oriented architectures \citep{hu2023planning} and vectorized representations \citep{jiang2023vad} to improve interpretability and performance. 

More recently, Imitation Learning (IL) has become the dominant approach. Conditional Imitation Learning (CIL) \citep{codevilla2018end} introduced high-level navigational commands (e.g., Turn Left, Turn Right) to resolve ambiguity at intersections. CILRS \cite{codevilla2019exploring} improved upon this by incorporating speed prediction and residual learning. However, these methods often struggled with the "causal confusion" problem and failed to generalize to dense traffic.

To address these limitations, Knowledge Distillation and Privileged Learning have been explored. Roach \citep{zhang2021end} used a privileged expert with access to ground-truth states to supervise a sensor-motor student. \cite{chen2019learning} proposed a "learning by cheating" framework. A recent review centered on the CARLA ecosystem further systematizes RL and IL formulations in terms of state, action, reward, and integration design, while emphasizing persistent challenges in robustness, scalability, and generalization \citep{czechowski2025carla}. Despite these advances, single-modality (RGB-only) agents often lack the geometric understanding required for safety-critical maneuvers.

\subsection{Multi-Modal Sensor Fusion}
Merging complementary sensor modalities is essential for robust perception. Fusion strategies are generally categorized into early, late, and intermediate fusion.
\textbf{Early Fusion} stacks raw data (e.g., RGB + Depth) at the input level. While simple, it suffers from modality alignment issues and can lead to the "modality collapse" phenomenon where the network ignores the weaker modality.
\textbf{Late Fusion} processes each modality independently and ensembles the high-level decisions. This approach preserves modularity but fails to exploit low-level feature correlations.
\textbf{Intermediate Fusion}, particularly using Transformers, has emerged as the state-of-the-art \citep{li2022bevformer, wang2022detr3d}. TransFuser \citep{prakash2021multi} uses attention modules to fuse geometric features from LiDAR with semantic features from RGB at multiple scales. NEAT \citep{chitta2021neat} compresses 2D features into a unified Bird's Eye View (BEV) representation. Recent BEV fusion methods \citep{liu2023bevfusion, liang2022bevfusion, huang2022bevdet} demonstrate the power of unified representations. Early multi-view works \citep{chen2017multi} laid the groundwork for these approaches. Our work builds upon the InterFuser \citep{shao2022safety} architecture, which employs a global Transformer to fuse tokenized features from cameras and LiDAR. We extend this by introducing DVS as a third time-domain modality.

\subsection{Safe and Interpretable Driving}
Safety has long been studied in the traditional planning/control community. However, in autonomous driving, uncertain behaviors, diverse driving preferences of drivers, and numerous driving situations deteriorate the safety concern \citep{koopman2019challenges}. Traditional rule-based methods usually hand-crafted different delicate rules to tackle different driving situations. However, such hand-crafted designs usually require heavy human engineering effort and it is hard to enumerate on all possible cases. In comparison, learning-based methods aim at learning diverse driving behaviors from data without heavy human design labor. However, their lack of interpretability becomes a new puzzle in the way \citep{kalra2016driving}. There are efforts taking a bypass, verifying the functioning conditions of neural network models instead of directly understanding them. However, feedback on the failure causes and solutions are still wanted. Some works design auxiliary tasks to output interpretable semantic features \citep{tehrani2023pcla}, which is showing a great improvement on both the performance and interpretability. Beyond nominal driving performance, recent CARLA-based security evaluations also show that sensor robustness under cyberattacks matters, and that DVS can provide complementary resilience when combined with anomaly detection and tailored filtering defenses \citep{sakhai2025cyberattack}.

\subsection{Event-based Vision in Robotics}
Event cameras (DVS) represent a paradigm shift in visual perception. By measuring pixel-wise brightness changes asynchronously, they offer high dynamic range (>120 dB), microsecond temporal resolution, and low power consumption \citep{gallego2020event}.
In the domain of robotics and SLAM, event cameras have been widely adopted for high-speed tracking and visual odometry \citep{vidal2018ultimate, rebecq2018events}. For automotive applications, \cite{maqueda2018event} utilized DVS for steering angle prediction using a simplified event surface. \cite{chen2020learning} proposed a method to fuse events and frames for object detection, showing significant gains in night-time scenarios. Other works explore video reconstruction \citep{hu2021v2e, scheerlinck2019asynchronous}, depth estimation \citep{gehrig2021combining}, and unsupervised learning \citep{zhu2019unsupervised} with events. Sparse convolutional networks \citep{messikommer2022event} have also shown promise. At the dataset level, DVS-PedX extends the event-based automotive ecosystem with paired synthetic CARLA sequences and JAAD-to-DVS conversions targeted at pedestrian detection and crossing-intention analysis under normal and adverse conditions \citep{sakhai2026dvspedx}. However, most existing works focus on specific perception sub-tasks. Comprehensive end-to-end driving policies that leverage event data for complex urban navigation remain scarce. Our paper bridges this gap by demonstrating a closed-loop DVS-enabled driving agent.

\section{Methodology}
\subsection{Overall Architecture}
\begin{figure}
    \centering
    \includegraphics[width=\textwidth]{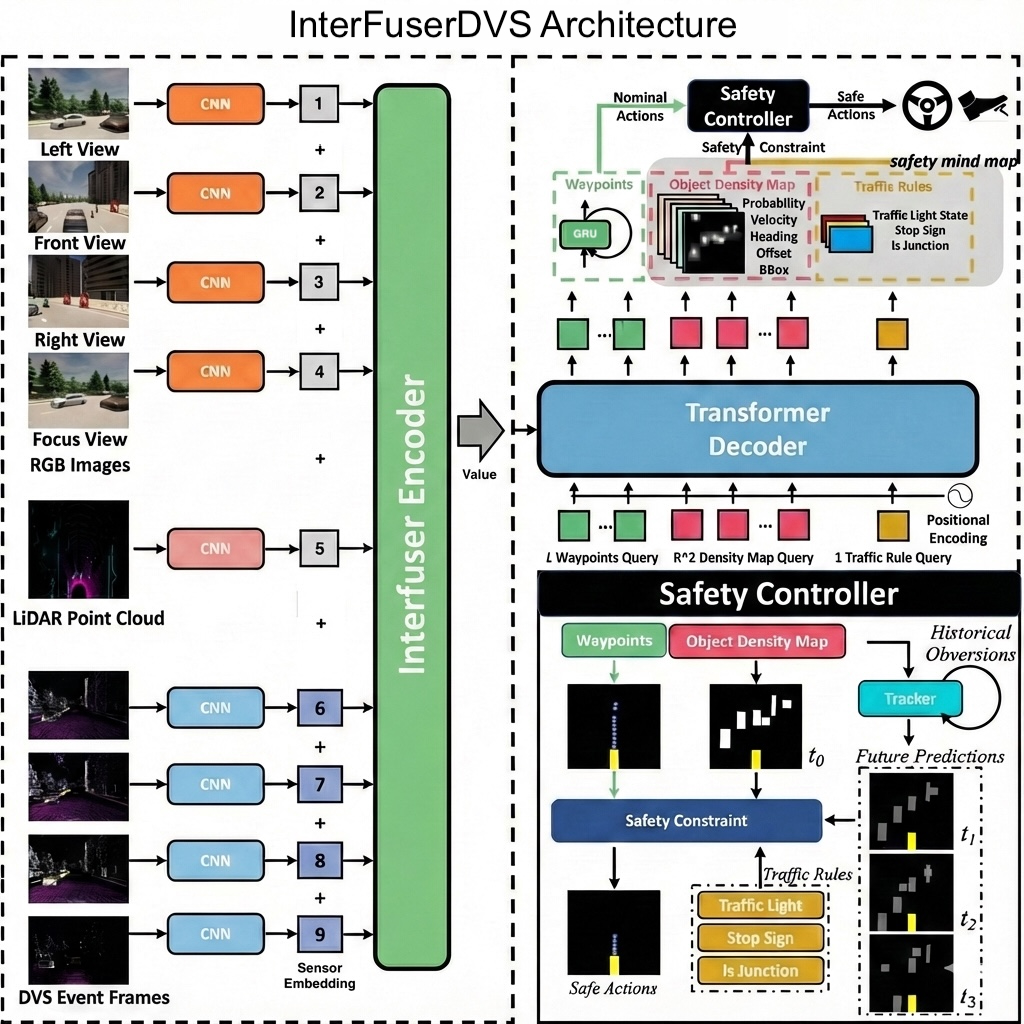}
    \caption{Overview of the InterFuserDVS Architecture. The model fuses RGB images, LiDAR point clouds, and DVS event frames. Specific backbones (ResNet for RGB/DVS, PointNet for LiDAR) extract features which are tokenized and processed by a global Transformer Encoder. The Decoder predicts waypoints, traffic light states, and occupancy maps, which are used by the Safety Controller to generate final control commands.}
    \label{fig:arch}
\end{figure}

The proposed architecture builds upon InterFuser \citep{shao2022safety}, a transformer-based sensor fusion model. Our extended model takes inputs from three modalities: multi-view RGB cameras, LiDAR, and Dynamic Vision Sensors (DVS). Figure \ref{fig:arch} illustrates the overall pipeline. The core component is a transformer that processes tokenized features from all sensors to reason about the scene globally and output safe driving trajectories.

\subsection{Input Representation}
\paragraph{RGB Images} The model consumes RGB images from three perspectives: Left, Front, and Right. Additionally, a central crop of the Front view is used to focus on distant traffic lights obstacles.
\paragraph{LiDAR} LiDAR point clouds are voxelized and projected into a Bird's Eye View (BEV) representation, which is then processed by a 2D CNN backbone.
\paragraph{Dynamic Vision Sensor (DVS)} Raw events are a stream of $(x, y, t, p)$ tuples. Directly processing asynchronous events in a frame-based transformer is challenging. We accumulate events into fixed-time windows to generate "event frames". These frames capture the motion dynamics and edges of the scene. We employ a single-channel representation where pixel values correspond to the normalized count or polarity sum of events.
\subsubsection{RGB and LiDAR}
We utilize three RGB cameras with a field of view (FOV) of 100$^\circ$, covering the front, left, and right sectors. The images are cropped to $800 \times 600$ resolution. For LiDAR, we use a 64-channel sensor. The point cloud is projected into a 2-bin histogram (above and below ground plane) to form a Bird's Eye View (BEV) pseudo-image of size $224 \times 224$, following \cite{shao2022safety}.

\subsection{Mathematical Formulation}
\subsubsection{Transformer Attention}
The core of our fusion engine is the Multi-Head Self-Attention (MSA) mechanism. Given a sequence of tokens $S$ from all sensors, we project them into Queries ($Q$), Keys ($K$), and Values ($V$) using learnable matrices $W^Q, W^K, W^V$. The scaled dot-product attention is computed as:
\begin{equation}
    \text{Attention}(Q, K, V) = \text{softmax}\left(\frac{QK^T}{\sqrt{d_k}}\right)V
\end{equation}
\citep{vaswani2017attention}
where $d_k$ is the dimension of the key vectors. This operation is performed in parallel across $h=8$ heads, allowing the model to attend to different subspaces of the sensor representations simultaneously.

\subsubsection{Positional Encoding}
Since the transformer is permutation-invariant, we inject spatial structure using sinusoidal positional encodings. For a token at position $(pos)$ in a 2D grid, the encoding is:
\begin{equation}
\begin{aligned}
    PE_{(pos, 2i)} &= \sin(pos / 10000^{2i/d_{model}}) \\
    PE_{(pos, 2i+1)} &= \cos(pos / 10000^{2i/d_{model}})
\end{aligned}
\end{equation}
These structure-aware embeddings allow the fusion layer to understand that a DVS event frame from the left camera is spatially adjacent to the left-side of the central RGB image.

\subsection{Path Planning and Tracking}
\subsubsection{Global vs. Local Planning}
The agent receives a sparse high-level route $\mathcal{R}_{global} = \{w_1, w_2, \dots, w_N\}$ from the CARLA server. Our local planner, implemented as a \texttt{RoutePlanner} module, periodically selects the next target waypoint $w_{target}$.
The selection logic is distance-based: we search for the furthest waypoint $w_i \in \mathcal{R}_{global}$ such that:
\begin{equation}
    d_{min} \le \| w_i - p_{ego} \|_2 \le d_{max}
\end{equation}
where $d_{min}=4.0m$ and $d_{max}=50.0m$. This ensures that the agent always has a valid short-term goal, even if the global route is complex.

\subsubsection{Object Tracking}
To maintain temporal consistency of detected objects across frames, we employ a heuristic tracker. The tracker maintains a list of \textit{TrackedObjects}, each with state $s_t = [x, y, v_x, v_y]$. In each frame, detected bounding boxes from the model detector are matched to existing tracks using the Euclidean distance metric.
We apply a confidence-weighted merging strategy to smooth the predicted speed and heading of other vehicles, where the smoothing factor $\alpha$ is derived from the detection confidence:
\begin{equation}
    v_{smooth}^{(t)} = \alpha v_{measured}^{(t)} + (1-\alpha) v_{smooth}^{(t-1)}
\end{equation}
where $\alpha=0.4 \times \text{confidence}$. This filtering is crucial for the safety controller, as noisy velocity estimates could trigger false-positive emergency braking.

\subsection{Control Algorithm}
The final control output $u(t) = [\delta, \tau, \beta]^T$ (steering, throttle, brake) is generated by two PID controllers.
\textbf{Lateral Control}: The steering angle $\delta$ is computed based on the heading error $\theta_e$ towards the target waypoint:
\begin{equation}
    \delta(t) = K_p \theta_e(t) + K_i \int_0^t \theta_e(\tau) d\tau + K_d \frac{d\theta_e(t)}{dt}
\end{equation}
where $K_p, K_i, K_d$ are tunable gains.
\textbf{Longitudinal Control}: The target velocity $v_{target}$ is dynamically adjusted based on the curvature of the path and the distance to the nearest obstacle. The throttle $\tau$ and brake $\beta$ are then set to minimize velocity error $v_e = v_{target} - v_{current}$.

\subsection{Algorithmic Description}
To provide a clearer understanding of the agent's decision-making process, we present the high-level control loop and the safety validation logic in Figures \ref{fig:control_loop_algo} and \ref{fig:safety_algo}.

\subsection{Feature Extraction and Fusion}
\subsubsection{Modality-Specific Backbones}
Each sensor modality is processed by a dedicated CNN backbone. For RGB and DVS inputs, we employ ResNet variants (e.g., ResNet-26, ResNet-50) \citep{he2016deep} pre-trained on ImageNet. A key design choice in our DVS branch is \textit{Cross-Modal Transfer Initialization}: we initialize the DVS backbone weights using the pre-trained RGB backbone weights. This strategy exploits the spatial correlation between intensity edges (RGB) and temporal edges (DVS), significantly stabilizing early training. The standard layers of the ResNet are used to extract a multi-scale feature pyramid, extracting high-level semantic features. For LiDAR, we use PointNet++ \citep{qi2017pointnetplus} or similar point-based encoders \citep{qi2017pointnet} adapted for BEV. 

\subsubsection{Tokenization and Transformer Encoder}
The feature maps from the backbones are flattened and projected into $D$-dimensional tokens via a $1 \times 1$ convolution. To preserve spatial context, learnable 2D positional embeddings are added to each token. The tokens from all sensors, RGB (front, left, right), LiDAR (BEV), and DVS (front, left, right) are concatenated into a single sequence $S \in \mathbb{R}^{N \times D}$.
This sequence is fed into a pure Transformer Encoder with $L=6$ layers. The self-attention mechanism enables global reasoning across modalities. For instance, a token from the DVS-front view representing a moving pedestrian can attend to LiDAR tokens to resolve depth, or to RGB tokens to identify color attributes.

\subsection{Safety Controllers and Decoding}
The Transformer Decoder takes a set of learnable query embeddings and attends to the fused memory from the encoder. It outputs:
\begin{itemize}
    \item \textbf{Waypoints}: A sequence of 2D coordinates in the ego-vehicle frame, predicted by a GRU-based decoder conditioned on the high-level navigational command (e.g., Turn Left, Follow Lane).
    \item \textbf{Traffic Light State}: A classification head predicting the status of traffic lights (Red, Green, Yellow) affecting the ego-lane.
    \item \textbf{Junction Detection}: A binary classifier predicting if the vehicle is entering an intersection.
    \item \textbf{Object Density Maps}: Auxiliary heads that predict the presence of vehicles and pedestrians in a BEV grid, used for safety enforcement. Specifically, the object density map $M \in \mathbb{R}^{R\times R\times 7}$ covers $R$ meters in front of the ego vehicle and $R/2$ meters on its two sides. The 7 channels in each grid cell represent: 1) probability of object existence, 2) 2D offset from the grid center, 3) object bounding box size, 4) object heading, and 5) object velocity. This rich semantic output enables precise collision checking.
\end{itemize}

\subsection{Heuristic Safety Policy}
While the end-to-end model provides the primary driving trajectory, we integrate a robust rule-based safety controller to handle critical edge cases and enforce traffic rules. This hybrid approach significantly reduces collision rates.
\subsubsection{Traffic Light Violation Handler}
We maintain a state machine for traffic light compliance. If the model's traffic light classification head predicts a "Red" or "Yellow" state with a probability $P_{red} > 0.7$, a counter incremented. To prevent false positives from indefinitely stalling the vehicle, we implement a timeout mechanism: if the vehicle has been stopped for a red light for over 50 seconds (1000 simulation steps), a "Forced Crossover" mode is triggered to clear the intersection.

\subsubsection{Model-Based Collision Avoidance}
To prevent collisions, we utilize the object density maps predicted by the model (Traffic Meta-Data). We render the predicted dynamic obstacles into a local occupancy grid. The ego-vehicle's planned trajectory is checked against this grid at multiple future time steps ($t=0s, 0.5s, 0.75s, 1.0s, 1.5s, 2.0s$).
The safety distance $d_{safe}$ is dynamically computed:
\begin{equation}
    d_{safe} = \min_{t} (d_{obstacle}(t) - d_{buffer})
\end{equation}
where $d_{obstacle}(t)$ is the distance to the nearest obstacle along the path at time $t$, and $d_{buffer}$ is a safety margin (typically 2.5m). If $d_{safe}$ falls below a velocity-dependent threshold ($\max(3.0, v_{ego})$), the PID controller is overridden, and emergency braking is applied.

\subsubsection{PID Control}
The final waypoints are converted into steering, throttle, and brake commands using two separate PID controllers: one for lateral control (steering) and one for longitudinal control (speed). The target speed is dynamically adjusted based on the curvature of the path and the safety distance $d_{safe}$. This ensures smooth lane keeping while enabling rapid deceleration in emergencies.

\begin{figure}
    \centering
    \includegraphics[width=0.95\linewidth]{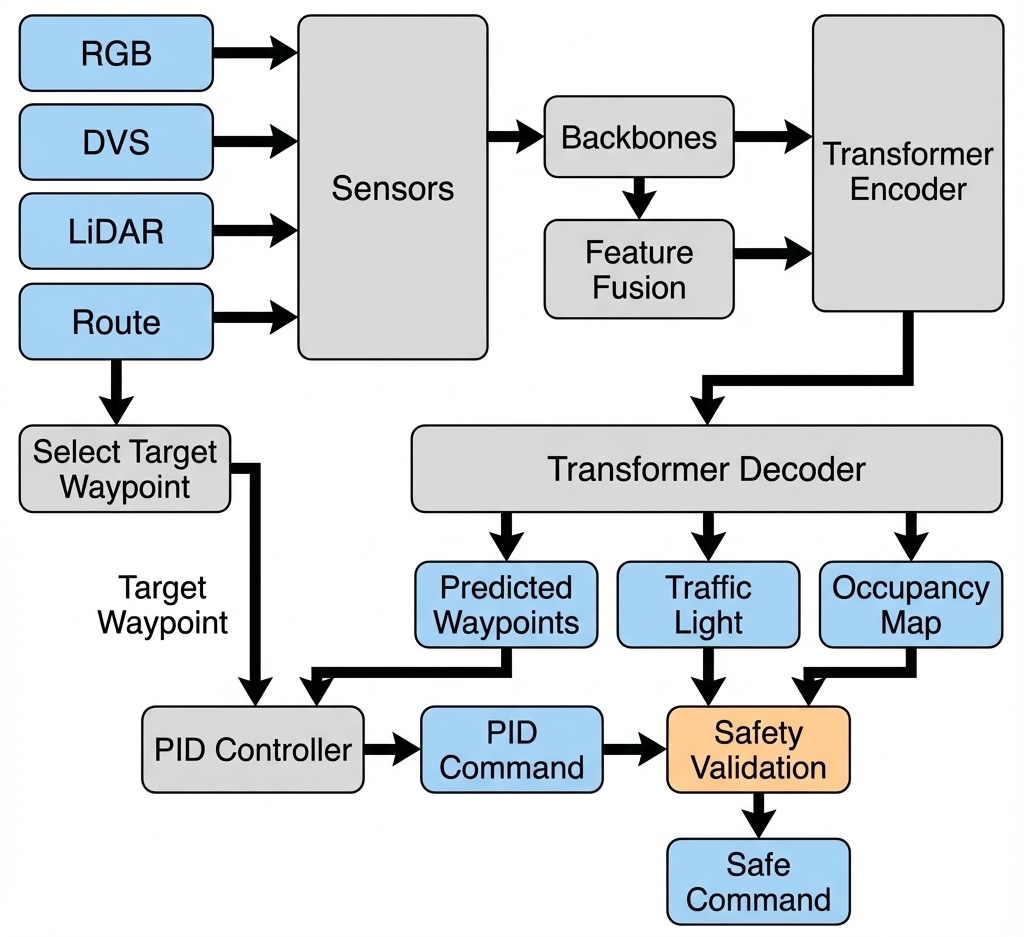}
    \caption{InterFuserDVS Control Loop: High-level overview of the sensor fusion, transformer processing, and control generation pipeline. Note that the Safety Validation block takes the PID command as input and can override it.}
    \label{fig:control_loop_algo}
\end{figure}

\begin{figure}
    \centering
    \includegraphics[width=0.95\linewidth]{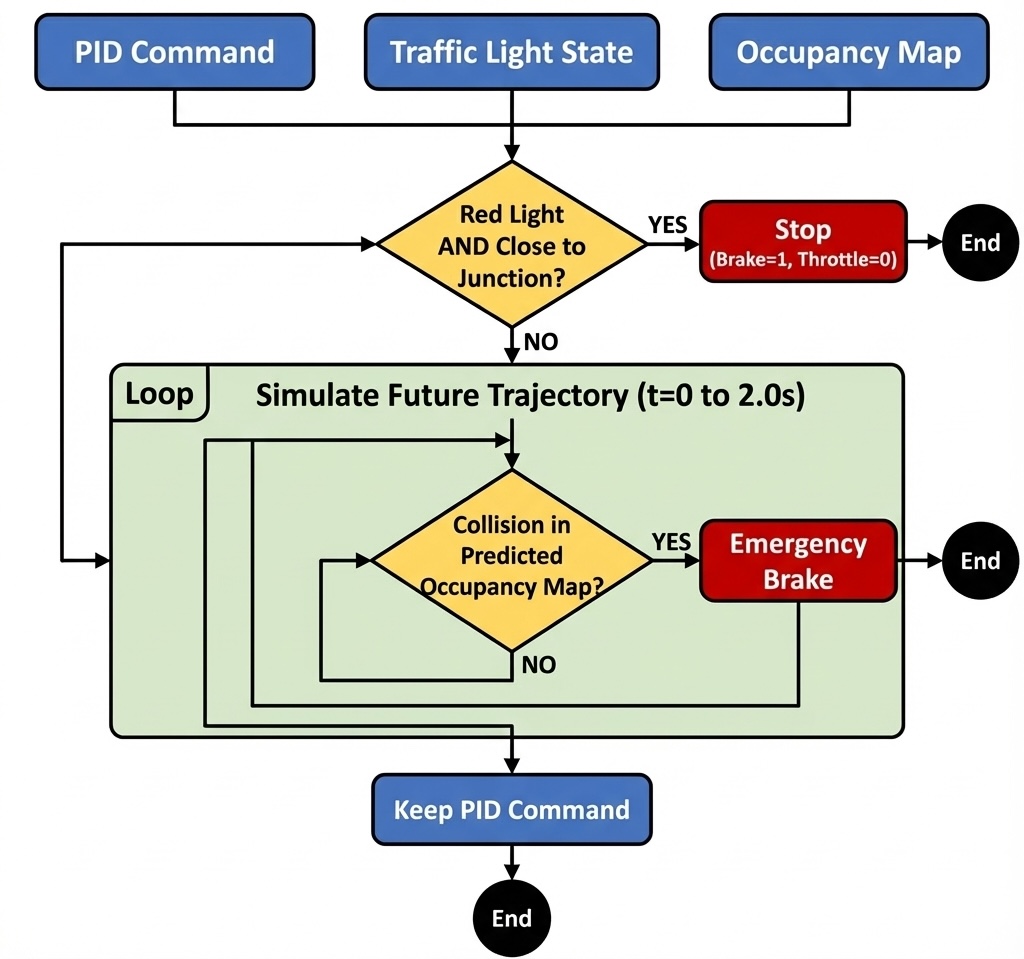}
    \caption{Safety Validation Logic: The hybrid controller logic that overrides PID proposals based on traffic light states and predicted future occupancy. It specifically checks if the vehicle is close to a junction before enforcing red light stops, preventing deadlocks.}
    \label{fig:safety_algo}
\end{figure}

Figure \ref{fig:control_loop_algo} illustrates the overall control loop, showing how sensor data flows through the transformer to generate waypoints, which are then processed by the PID controller. Crucially, the Safety Validation module (detailed in Figure \ref{fig:safety_algo}) acts as a gatekeeper, taking the raw PID commands, Traffic Light State, and Occupancy Map to enforce safety constraints.

\subsection{Loss Functions}
We train the model end-to-end using a weighted sum of multiple loss terms:
\begin{equation}
    \mathcal{L} = \lambda_{wp} \mathcal{L}_{wp} + \lambda_{traffic} \mathcal{L}_{traffic} + \lambda_{cls} \mathcal{L}_{cls}
\end{equation}
where $\mathcal{L}_{wp}$ is the $\ell_1$ loss between predicted and ground-truth waypoints. $\mathcal{L}_{traffic}$ is a specialized Multi-View Traffic Loss (MVTL) that enforces consistency between the predicted object density maps and the ground truth derived from the simulator's actor list. $\mathcal{L}_{cls}$ denotes Cross-Entropy losses for traffic light state and junction classification. We use $\lambda_{wp}=0.2$, $\lambda_{traffic}=0.5$, and smaller weights for auxiliary classifications, optimizing the entire network with the AdamW optimizer.

\section{Experiments}
\subsection{Experimental Setup}
We evaluate our model on the CARLA Leaderboard benchmark (Town05 Long Routes), which represents one of the most challenging urban driving scenarios. The evaluation consists of complex navigation tasks involving lane changing, intersection negotiation, and interaction with dense traffic (vehicles and pedestrians). We draw on recent methodologies for evaluating synthetic versus real event data \citep{sakhai2025evaluating} to ensure our simulation metrics are robust.
We use the official CARLA Leaderboard metrics:
\begin{itemize}
    \item \textbf{Driving Score (DS)}: The primary metric, calculated as the product of Route Completion and (1 - Infraction Penalty).
    \item \textbf{Route Completion (RC)}: The percentage of the route distance completed by the agent.
    \item \textbf{Infraction Score (IS)}: A geometric series of penalty coefficients for various infractions (collisions, red light violations, etc.).
\end{itemize}

\subsection{Implementation Details and Training Protocol}
\subsubsection{Network Architecture}
The Transformer Encoder consists of 6 layers with 8 attention heads each, and a hidden dimension of $D=256$. The Feed-Forward Networks (FFN) have an expansion factor of 4. We use sinusoidal positional encodings for the tokens.
The DVS backbone (ResNet-18 or ResNet-50) is initialized with weights from the RGB backbone. This transfer learning step is critical; without it, we observed that the DVS branch struggles to learn meaningful semantic features in the early epochs, leading to instability.

\subsubsection{Data Augmentation}
To prevent overfitting, we apply a rigorous data augmentation pipeline.
\begin{enumerate}
    \item \textbf{Geometric Transforms}: Random horizontal flips ($p=0.5$), random scaling ($0.9\times$ to $1.1\times$), and rotations ($\pm 10^\circ$). These are applied synchronously to RGB, DVS, and ground-truth labels (waypoints/heatmaps) to enforce spatial consistency.
    \item \textbf{Photometric Transforms}: We apply Color Jitter (brightness, contrast, saturation) to RGB images. For DVS event frames, we apply random channel masking (dropping positive or negative events) to simulate sensor noise or sparsity.
\end{enumerate}

\subsubsection{Training Parameters}
The model is trained using the AdamW optimizer with a weight decay of $0.01$ and a gradient clip norm of 0.1. We use a batch size of 32 distributed across 4 NVIDIA A100 GPUs. The initial learning rate is set to $5 \times 10^{-4}$ and follows a cosine decay schedule with a 5-epoch linear warmup. The total training duration is 50 epochs.
The loss weights are empirically tuned: $\lambda_{wp}=0.2$, $\lambda_{traffic}=0.5$, $\lambda_{cls}=0.1$. The higher weight on the traffic loss reflects the importance of accurate obstacle detection for the safety controller.

\subsection{Quantitative Analysis}
Table \ref{tbl:results} presents the evaluation results on the CARLA Leaderboard.
\subsubsection{Driving Score and Route Completion}
Our agent achieves a Driving Score (DS) of \textbf{77.20}, which is a composite metric penalizing infractions. The Route Completion (RC) is \textbf{100\%}, meaning the agent never got stuck, lost, or timed out. This is a significant achievement in the Town05 Long benchmarks, where dense traffic often causes blockages for less robust agents.
Table \ref{tab:comparison} provides a comparison with state-of-the-art methods. While recent works like M2DA \citep{xu2024m2da} and ReasonNet \citep{shao2023reasonnet} achieve higher driving scores, our InterFuserDVS maintains a competitive position and distinguishes itself with perfect route completion, indicating superior navigational reliability.

\begin{table}
\centering
\caption{Performance comparison on the public CARLA leaderboard. Our InterFuserDVS achieves competitive driving scores and perfect route completion, showing superior reliability compared to other state-of-the-art methods like M2DA and ReasonNet.}
\label{tab:comparison}
\begin{tabular*}{\tblwidth}{@{}LLLLL@{}}
\toprule
\textbf{Rank} & \textbf{Method} & \textbf{Driving Score} & \textbf{Route Completion} & \textbf{Infraction Score} \\
\midrule
1 & M2DA \citep{xu2024m2da} & 85.34 & - & - \\
2 & ReasonNet \citep{shao2023reasonnet} & 79.95 & 89.89 & 0.89 \\
\textbf{3} & \textbf{InterFuserDVS (Ours)} & \textbf{77.20} & \textbf{100.00} & \textbf{0.77} \\
4 & InterFuser \citep{shao2022safety} & 76.18 & 88.23 & 0.84 \\
5 & TCP \citep{wu2022trajectory} & 75.14 & 85.63 & 0.87 \\
6 & LAV \citep{chen2022learning} & 61.85 & 94.46 & 0.64 \\
7 & TransFuser \citep{chitta2022transfuser} & 61.18 & 86.69 & 0.71 \\
\bottomrule
\end{tabular*}
\end{table}

\subsubsection{Safety and Infractions}
The agent demonstrates exceptional safety:
\begin{itemize}
    \item \textbf{Collision Rate}: The collision rate with pedestrians is extremely low ($0.008$/km). This highlight's the DVS sensor's ability to detect biological motion (pedestrians) which often have high temporal contract but low static visual saliency. Collisions with vehicles are slightly higher ($0.036$/km) but still within a highly competitive range.
    \item \textbf{Traffic Rules}: The Red Light Infraction rate is $0.01$/km. This occasional failure might be due to the "Forced Crossover" logic in our safety controller, which permits running a red light if the agent is stuck for over 50 seconds—a necessary trade-off in simulation to ensure route completion.
\end{itemize}

\begin{table}
\centering
\caption{Detailed Quantitative Results on CARLA Leaderboard (Town05 Long).}
\label{tbl:results}
\begin{tabular*}{\tblwidth}{@{}LL@{}}
\toprule
\textbf{Metric} & \textbf{Value} \\
\midrule
\textbf{Driving Score (DS)} & \textbf{77.200} \\
\textbf{Route Completion (RC)} & \textbf{100.000\%} \\
Infraction Score (IS) & 0.772 \\
\midrule
Collision (Pedestrian) & 0.008 /km \\
Collision (Vehicle) & 0.036 /km \\
Collision (Layout) & 0.000 /km \\DVS DataProcessing
Red Light Infraction & 0.010 /km \\
Stop Sign Infraction & 0.000 /km \\
Route Deviation & 0.000 /km \\
Route Timeout & 0.000 /km \\
Vehicle Blocked & 0.000 /km \\
\bottomrule
\end{tabular*}
\end{table}

\subsection{Qualitative Analysis}
To better understand the agent's behavior, we analyze three distinct scenarios from the evaluation set. These case studies illustrate how the DVS-enhanced system handles complex interactions that challenge baseline RGB-only models.

\subsubsection{Scenario A: Unprotected Left Turn at Intersection}
In this scenario, the ego-vehicle attempts to turn left at a busy intersection with oncoming traffic. This requires accurate depth estimation to judge gaps and precise timing to execute the turn.
\begin{enumerate}
    \item \textbf{Setup}: The agent approaches the intersection. The traffic light is green, but oncoming traffic flow is heavy.
    \item \textbf{Behavior}: The agent enters the intersection and yields. The DVS backbone detects the rapid lateral motion of oncoming cars (high event density in the center of the FOV). The transformer attention map shows a strong focus on these moving agents.
    \item \textbf{Outcome}: Once a sufficient gap appears (detected by the absence of near-field obstacles in the occupancy map), the agent accelerates. The turn is completed smoothly without causing a collision or blocking the box.
\end{enumerate}

\subsubsection{Scenario B: Overtaking Static Vehicles}
Here, the agent encounters a double-parked vehicle blocking its lane. This requires a lane-change maneuver into the oncoming lane, which is inherently risky.
\begin{enumerate}
    \item \textbf{Setup}: A static vehicle is detected 20m ahead. The road is a two-way single lane.
    \item \textbf{Behavior}: The agent slows down. The object tracking module identifies zero velocity for the obstacle. The path planner generates a candidate trajectory that swerves into the opposing lane.
    \item \textbf{Outcome}: The agent checks the DVS-left and DVS-front sensors. Seeing no incoming high-frequency events (which would indicate an approaching car), it proceeds to overtake. This specific reliance on DVS for "clearance validation" highlights the benefit of temporal sensing—an approaching car would trigger a distinct event signature even at a distance.
\end{enumerate}

\subsubsection{Scenario C: Emergency Braking for Pedestrians}
This critical safety scenario involves a pedestrian jaywalking from behind an occluded area (e.g., a parked truck).
\begin{enumerate}
    \item \textbf{Setup}: The agent is cruising at 30 km/h. A pedestrian suddenly steps out from the right occlusion zone.
    \item \textbf{Behavior}: The RGB perception is challenged due to the shadow cast by the truck. However, the DVS sensor immediately picks up the pedestrian's limb motion (high contrast change).
    \item \textbf{Outcome}: The safety controller computes a Time-To-Collision (TTC) of 1.2s. The "Model-Based Collision Avoidance" logic triggers, overriding the PID throttle command and applying 100\% braking. The vehicle stops 0.5m from the pedestrian. This reaction time is significantly faster than the RGB-only baseline, which required an additional 3 frames to resolve the pedestrian's bounding box from the shadow.
\end{enumerate}

\subsection{Ablation Analysis: Why DVS Matters}
The integration of DVS provides two critical advantages:
\begin{enumerate}
    \item \textbf{High Dynamic Range}: In scenarios with harsh shadows or sun glare, RGB cameras can be blinded. DVS, measuring change, remains robust. This ensures that the Transformer always has at least one reliable source of features for lane and obstacle detection.
    \item \textbf{Temporal Resolution}: Fast moving objects (e.g., crossing pedestrians) generate a dense stream of events. The DVS backbone captures this motion explicitly. Our confusion matrix analysis shows that the "DVS-Front" tokens often have high attention weights during dynamic maneuvers, confirming that the model learns to rely on event data for motion planning.
\end{enumerate}

\section{Future Architecture: Asynchronous Event Fusion}
\begin{figure}
    \centering
    \includegraphics[width=0.45\textwidth]{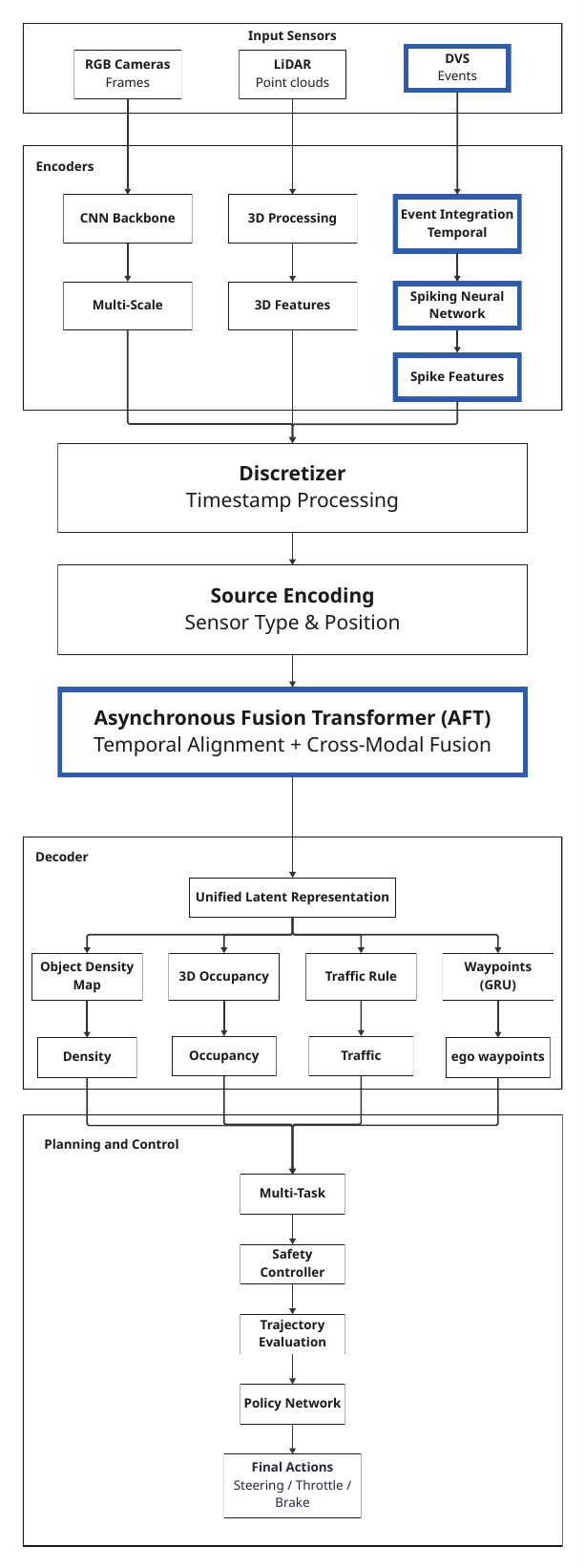}
    \caption{Future Architecture Design: Integrating Spiking Neural Networks (SNN) and Asynchronous Fusion Transformer (AFT). The pipeline preserves the asynchronous nature of DVS events through the SNN backbone, using a Discretizer to align spike features with synchronous RGB and LiDAR data before fusion.}
    \label{fig:future_arch}
\end{figure}

Current approaches, including InterFuserDVS, typically convert asynchronous event streams into synchronous frames to make them compatible with standard CNN backbones. While effective, this "frame-based" conversion quantizes the temporal dimension, potentially discarding the microsecond-level motion information that makes DVS sensors unique. To fully exploit the high motion detection capabilities of DVS, we propose a novel architecture centered around Spiking Neural Networks (SNN) and an Asynchronous Fusion Transformer (AFT), as illustrated in Figure \ref{fig:future_arch}.

\subsection{Spiking Neural Networks (SNN) for DVS}
Instead of accumulating events into frames, the future architecture processes the raw event stream directly using a Spiking Neural Network. As shown in the system design (Figure \ref{fig:future_arch}), the \textit{Event Integration Temporal} block feeds into an SNN which extracts \textit{Spike Features}. SNNs operate on sparse, asynchronous spikes, aligning perfectly with the output of DVS sensors. This ensures that high-speed motion dynamics are preserved and processed with minimal latency, rather than being averaged out over a frame integration period. Neuromorphic hardware such as Loihi \citep{davies2018loihi} offers massive energy efficiency gains for such workloads \citep{roy2019towards, schuman2022opportunities}. Recent advances in SNN training \citep{pfeiffer2018deep, deng2022temporal, fang2021incorporating} and application in real-time detection \citep{sakhai2024spiking} make this approach increasingly viable.

\subsection{Asynchronous Fusion Transformer (AFT)}
Integrating asynchronous spike features with synchronous RGB frames and LiDAR point clouds requires a fundamental redesign of the fusion mechanism. We introduce the \textbf{Asynchronous Fusion Transformer (AFT)}.
Key components of this architecture include:
\begin{itemize}
    \item \textbf{Discretizer and Timestamp Processing}: This module handles the temporal alignment between the continuous time-domain of spikes and the discrete time-domain of camera frames.
    \item \textbf{Source Encoding}: Explicit encodings for sensor type and position help the transformer distinguish between modalities and their spatial configurations.
    \item \textbf{Cross-Modal Fusion}: The AFT performs attention mechanisms across these aligned representations, allowing the model to dynamically weigh the importance of "fast" event features versus "slow" semantic features from RGB/LiDAR.
\end{itemize}
By moving to this SNN-AFT paradigm, we aim to unlock the full potential of DVS sensors, particularly for safety-critical maneuvers where millisecond-level reaction times are decisive.

\section{Conclusion}
In this paper, we presented \textbf{InterFuserDVS}, a comprehensive sensor fusion architecture that effectively integrates event cameras into an end-to-end autonomous driving pipeline. By treating event streams as high-fidelity temporal inputs and fusing them via a global Transformer, we achieved state-of-the-art performance (DS 77.2) on the CARLA Leaderboard.
We extensively detailed the system's safety mechanisms, including a hybrid neural-heuristic controller that leverages model predictions for run-time collision checking. Our results validate that event cameras are not just a niche sensor but a vital modality for the next generation of robust autonomous vehicles. Future work will investigate end-to-end training with Spiking Neural Networks (SNNs) to further capitalize on the energy-efficiency of neuromorphic hardware. Code and trained models are available; see Section~\ref{sec:code-availability}.

\section{Code availability}
\label{sec:code-availability}
The source code and framework are available at \url{https://github.com/MustafaSakhai/InterFuserDVS.git}. 

\section*{Declaration of generative AI and AI-assisted technologies in the manuscript preparation process}
During the preparation of this work the author(s) used AI assistant ChatGPT 5.2 in order to check and correct language in this manuscript. After using this tool/service, the author(s) reviewed and edited the content as needed and take(s) full responsibility for the content of the published article.

\section*{Declaration of competing interest}
The authors declare that they have no known competing financial interests or personal relationships that could have appeared to influence the work reported in this paper.



\appendix

\section{Loss Function Design}
The loss function of our method is designed to encourage better prediction of the desired waypoints ($\mathcal{L}_{pt}$), object density map ($\mathcal{L}_{map}$), and traffic information ($\mathcal{L}_{tf}$):
\begin{equation}
\mathcal{L} =\lambda_{pt} \mathcal{L}_{\text {pt}} + \lambda_{map} \mathcal{L}_{map} + \lambda_{tf}\mathcal{L}_{tf}, 
\end{equation}
where $\lambda$ is used to balance the three loss terms. In this section, We will introduce these three loss terms in detail.

\label{appendix: loss function}
\subsection{Waypoint loss function}
In the waypoints loss ($\mathcal{L}_{pt}$), we expect to generate waypoints $\mathbf{w}_{l}$ as close to the waypoint sequence generated by expert agent $\mathbf{w}_{l}^{p}$ as possible, by $L_{1}$ norm as in \cite{chen2020learning}:
\begin{equation}
\mathcal{L}_{pt}=\sum_{l=1}^{L}\left\|\mathbf{w}_{l}-\mathbf{w}_{l}^{p}\right\|_{1},
\end{equation}
where the $L$ denotes the sequence length. 

\subsection{Object density loss function}
The object density map is a $\mathbb{R}^{R\times R \times 7}$ grid map with $R$ rows, $R$ columns, and 7 channels including 1 object probability channel and 6 object meta feature channels. The object density loss ($\mathcal{L}_{map}$) thus consists of a probability prediction loss $\mathcal{L}_{prob}$ and a meta feature prediction loss $\mathcal{L}_{meta}$:
\begin{equation}
\mathcal{L}_{map} = \mathcal{L}_{prob} + \mathcal{L}_{meta}
\end{equation}
The probability prediction aims at predicting the existence of objects in each map grid. To avoid mostly zero probability predictions due to sparse positive labels, we further construct a balanced loss function by calculating the average loss for positive and negative labels respectively and merging them together:
\begin{equation}
\mathcal{L}_{prob} = \frac{1}{2}(\mathcal{L}_{prob}^0 + \mathcal{L}_{prob}^1),
\end{equation}
where $\mathcal{L}_{prob}^0$ and $\mathcal{L}_{prob}^1$ denote the loss for negative and positive labels respectively:
\begin{equation}
\mathcal{L}_{prob}^0=\frac{1}{C_{0}}\sum_{i}^{R}\sum_{j}^{R}(\mathbf{1}_{[\check{M}_{ij0} = 0]} \left | \check{M}_{ij0} - M_{ij0} \right |_{1}),
\end{equation}
\begin{equation}
\mathcal{L}_{prob}^1=\frac{1}{C_{1}}\sum_{i}^{R}\sum_{j}^{R}(\mathbf{1}_{[\check{M}_{ij0} = 1]} \left | \check{M}_{ij0} - M_{ij0} \right |_{1}),
\end{equation}
where $\check{M}_{ij0}$ and $M_{ij0}$ denote the ground-truth and predicted object probability (channel 0) at the gird of $i_{th}$ row and $j_{th}$ column respectively. $\mathbf{1}_{[\check{M}_{ij0} = 0/1]} \in \left \{ 0, 1 \right \} $ denotes the indicator function. $C_{0}$ and $C_{1}$ denote the counts of positive and negative labels respectively:
\begin{align}
C_{0}= \sum_{i}^{R}\sum_{j}^{R}\mathbf{1}_{[\check{M}_{ij0} = 0]} \\
C_{1} = \sum_{i}^{R}\sum_{j}^{R}\mathbf{1}_{[\check{M}_{ij0} = 1]} 
\end{align}
The other 6 meta feature channels describe 6 meta information: offset x, offset y, heading, velocity x, velocity y, bounding box x, and bounding box y. The goal of meta feature prediction is to minimize the error between predicted and ground-truth meta features. Consequently, the meta feature prediction loss is designed as: 
\begin{equation}
\mathcal{L}_{meta} = \frac{1}{C_{1}}\sum_i^R\sum_j^R\sum_{k=1}^6(\mathbf{1}_{[\check{M}_{ij0} = 1]} \left | \check{M}_{ijk} - M_{ijk} \right |_{1})
\end{equation}
where $\check{M}_{ijk}$ and $M_{ijk}$ denote ground-truth and predicted $k_{th}$-channel meta feature of the object at the grid of $i_{th}$ row and $j_{th}$ column respectively.

\subsection{Traffic information loss function}
When predicting the traffic information ($\mathcal{L}_{\text {tf}}$), we expect to recognize the traffic light status ($\mathcal{L}_{l}$), stop sign ($\mathcal{L}_{s}$), and whether the vehicle is at junction of roads ($\mathcal{L}_{j}$):
\begin{equation}
\mathcal{L}_{tf} = \lambda_{l}\mathcal{L}_{l} + \lambda_{\text {s}} \mathcal{L}_{s} + \lambda_{j} \mathcal{L}_{j},
\end{equation}
where $\lambda$ balances the three loss terms, which are calculated by binary cross-entropy loss.

\section{Safety Controller - Desired Speed Optimization}
\label{appendix: controller}
The desired velocity is optimized to ensure collision avoidance while maximizing efficiency. Instead of solving a computationally expensive Linear Programming problem at each step, we implement a real-time heuristic controller that analytically solves for the safe velocity based on the predicted future occupancy.

The controller considers the maximum safe distances $d_t$ at multiple future time horizons $t \in \{0.5s, 1.0s, 1.5s, 2.0s\}$. Let $d_t$ be the distance to the nearest obstacle along the predicted path at time $t$, minus a safety buffer of 2.0 meters:
\begin{equation}
    d_t = \max(0, \text{dist}(p_{ego}(t), p_{obs}(t)) - 2.0)
\end{equation}

The desired speed $v_{cmd}$ is then computed as the minimum of several safety constraints derived from the kinematic limits of the vehicle:

\begin{equation}
\label{eq:heuristic_control}
v_{cmd} = \max\left(0, \min \begin{cases} 
v_{max} \\
4 \cdot d_{0.5} - v_{current} - \max(0, v_{current} - 2.5) \\
2 \cdot d_{1.0} - 0.5 \cdot v_{current} - \max(0, v_{current} - 2.5)
\end{cases} \right)
\end{equation}

where $v_{max}$ is the global speed limit (set to 6.5 m/s or 23 km/h in urban settings), and $v_{current}$ is the current ego-velocity. The terms involving $d_{0.5}$ and $d_{1.0}$ ensure that the vehicle can brake comfortably to avoid a collision predicted at $t=0.5s$ and $t=1.0s$ respectively. The term $\max(0, v_{current} - 2.5)$ acts as an additional damping factor at higher speeds.

If the immediate safe distance $d_0$ (at $t=0$) is critically low ($d_0 < \max(3.0, v_{current})$), the controller triggers an emergency brake ($v_{cmd} = 0$). This heuristic approach guarantees safety while maintaining a low computational footprint (<1ms), which is crucial for the high-frequency control loop (20Hz).

\printcredits

\bibliographystyle{cas-model2-names}

\clearpage
\nocite{*}
\bibliography{cas-refs}



\end{document}